\documentclass[10pt,twocolumn,letterpaper]{article}

\usepackage{cvpr}              

\usepackage[dvipsnames]{xcolor}

\definecolor{cvprblue}{rgb}{0.21,0.49,0.74}
\usepackage[pagebackref,breaklinks,colorlinks,citecolor=cvprblue]{hyperref}
\usepackage{textcomp}
\usepackage{gensymb}
\usepackage{multicol}
\usepackage{cuted}
\usepackage{caption}
\usepackage{ragged2e} 
\def\paperID{281} 
\def\confName{3DV\xspace}
\def\confYear{2025\xspace}

\title{CFPNet: Improving Lightweight ToF Depth Completion via Cross-zone Feature Propagation}

\author{
\begin{tabular}{cc}
Laiyan Ding & Rui Huang* \\
{\small\tt laiyanding@link.cuhk.edu.cn} & {\small\tt ruihuang@cuhk.edu.cn} \\[1em]
Hualie Jiang & Rui Xu \\
{\small\tt jianghualie@insta360.com} & {\small\tt xurui@insta360.com}
\end{tabular}
}

\begin{document}
\maketitle
\begin{abstract}
Depth completion using lightweight time-of-flight (ToF) depth sensors is attractive due to their low cost. However, lightweight ToF sensors usually have a limited field of view (FOV) compared with cameras. Thus, only pixels in the zone area of the image can be associated with depth signals. Previous methods fail to propagate depth features from the zone area to the outside-zone area effectively, thus suffering from degraded depth completion performance outside the zone. To this end, this paper proposes the CFPNet to achieve cross-zone feature propagation from the zone area to the outside-zone area with two novel modules. The first is a direct-attention-based propagation module (DAPM), which enforces direct cross-zone feature acquisition. The second is a large-kernel-based propagation module (LKPM), which realizes cross-zone feature propagation by utilizing convolution layers with kernel sizes up to 31. CFPNet achieves state-of-the-art (SOTA) depth completion performance by combining these two modules properly, as verified by extensive experimental results on the ZJU-L5 dataset. The code is available at \url{https://github.com/denyingmxd/CFPNet}.
\end{abstract}    
\section{Introduction}
\label{sec:intro}

Depth completion from sparse depth measurements is an essential component of many tasks, including SLAM \cite{liu2023multi}, novel view synthesis \cite{wang2023digging}, and robot navigation \cite{ma2019sparse}. Previous depth completion methods usually simulate sparse depth inputs randomly from depth maps acquired by RGB-D cameras, e.g., RealSense. Nevertheless, it is impractical to obtain such sparse depth as inputs for real applications. Recently, ToF sensors have been applied to depth super-resolution \cite{he2021towards} and depth completion \cite{li2022deltar} due to their low power consumption and cost-effectiveness compared with RGB-D cameras. In this paper, we focus on the depth completion task, particularly implemented on a popular type of lightweight ToF sensor (e.g., ST VL53L5CX \cite{VL53L5CX}, denoted as L5), though our method is not limited to it. Despite its low resolution (e.g., $8 \times 8$), the low power consumption (e.g., 200mW) and low cost (e.g., $\$ 6$) of L5 allows it to be deployed in more applications. Note that we refer to this depth estimation task assisted by a lightweight ToF sensor as depth completion due to their similarity in input depth signals \cite{sun2022mipi}.

\begin{figure}[t]
    \centering
    \includegraphics[width=0.5\textwidth]{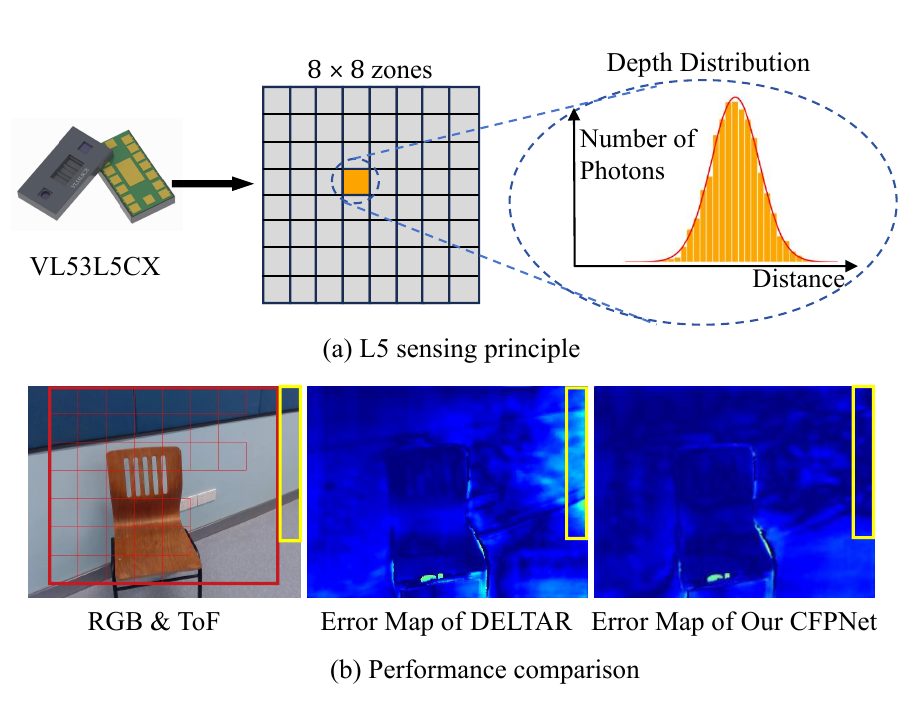}
    \caption{L5 sensing principle and performance comparison. (a) L5 would return zones of resolution $8 \times 8$, and each zone provides depth distribution information. (b) We overlay aligned zone areas on the paired RGB image and display the error maps of DELTAR and our CFPNet. The largest rectangle is the zone area, and some zones are missing due to too few received photons or inconsistency in measurement. Notice that CFPNet obtains smaller errors in outside-zone areas (the yellow rectangle). } 
    \label{fig:motivation}
\end{figure}

To get a preliminary overview of depth completion with a lightweight ToF sensor, we illustrate the sensing principle of L5 in Fig. \ref{fig:motivation} (a) as an example. Unlike conventional ToF sensors, L5 is lightweight and has extremely low resolution (e.g., $8 \times 8$). Each zone in L5 produces a depth distribution about the corresponding 3D scene by counting the number of photons returned in discretized time intervals. DELTAR fits this distribution with a Gaussian distribution and transmits the mean and variance to save bandwidth and energy \cite{li2022deltar}. In Fig. \ref{fig:motivation} (b), we overlay the zones on the image. The zone area is the largest red rectangle, and each small red rectangle corresponds to one ToF measurement. Note that some zones are missing due to too few received photons or inconsistency in measurement.  

Though depth inputs collected from lightweight ToF sensors (e.g., L5) are sparse and noisy, researchers have made depth completion from them plausible. DELTAR is a RGB-guided lightweight-ToF-based depth completion model \cite{li2022deltar}. At the core of DELTAR, an effective D-to-image module conducts feature fusion between depth and RGB features patchwisely in the zone area. Then, a self-attention layer \cite{vaswani2017attention} is leveraged to propagate the fused features to outside-zone areas. However, since this self-attention \cite{vaswani2017attention} queries values based on similarities, pixels in outside-zone areas may gather more information from outside-zone pixels which do not contain ToF information than from in-zone pixels. Thus, it is unlikely to propagate depth features effectively from zone areas to outside-zone areas. To tackle the restricted cross-zone feature propagation issue, we introduce a direct-attention-based module and a large-kernel-based module to establish steady and long-range dependencies between the two areas.

\textbf{Direct-Attention-based Propagation Module.} DAPM is based on the attention mechanism \cite{vaswani2017attention}, which enables feature propagation regardless of pixel distances in the image. Specifically, we directly perform cross-attention from in-zone pixels to outside-zone pixels. Thus, outside-zone pixels can query information from in-zone pixels dynamically. More importantly, our DAPM avoids feature acquisition from outside-zone areas where ToF information does not exist during feature propagation.

\textbf{Large-Kernel-based Propagation Module.} LKPM incorporates a convolution layer of a large kernel size (e.g., $31 \times 31$) \cite{ding2022scaling}. Large-kernel CNNs have been shown to have larger effective receptive fields \cite{luo2016understanding} compared with small-kernel CNNs. Consequently, we use convolution layers of large kernels to establish long-range dependencies between pixels from zone areas and outside-zone areas. Moreover, the interaction between input signals in convolution depends on location rather than similarities as in attention \cite{bello2019attention}. Thus, with the large-kernel design, LKPM is not likely to fall into the situation that feature acquisition only comes from outside-zone areas, mitigating potential limitations.

Owing to the proposed DAPM and LKPM, our CFPNet can reduce the errors in outside-zone areas as in Fig. \ref{fig:motivation} (b). As a result, compared with the previous method \cite{li2022deltar}, we reduce the mean absolute relative error (REL) from 0.127 to 0.103 on the ZJU-L5 dataset\cite{li2022deltar}. Notably, we decrease REL and RMSE by 46.5\% and 30.8\%, respectively, when the ToF is of resolution $2 \times 2$. In summary, our contributions are as follows:

\begin{enumerate}
    \item We notice that outside-zone areas suffer from a great performance drop compared with in-zone areas. Furthermore, we propose CFPNet to alleviate this degradation with more effective cross-zone feature propagation.
    \item Our CFPNet contains two feature propagation modules, namely DAPM and LKPM. DAPM allows direct feature propagation with the help of cross-attention, and LKPM propagates features using convolution layers of large kernel sizes.
    \item CFPNet achieves remarkable performance gain on the ZJU-L5 dataset over previous methods. Codes will be released for peer research.
\end{enumerate}

\section{Related Work}
\label{sec:related_work}

\subsection{Depth Completion} Early works of depth completion aim to predict pixel-wise depth maps given sparse depth \cite{uhrig2017sparsity} without RGB guidance. Yet, RGB-guided approaches generally outperform unguided ones \cite{hu2022deep}. Thus, methods that take both RGB and sparse depth as inputs have been proposed. CSPN \cite{cheng2019learning} refine depth prediction within a local convolutional context. Nevertheless, fixed neighbours defined by local windows prevent information propagation in larger contexts. Consequently, NLSPN \cite{park2020non} allow depth refinement with non-local neighbours, CSPN$++$ \cite{cheng2020cspn++} employ convolutions contexts of different kernel sizes, and PENet \cite{hu2021penet} introduce dilated CSPN$++$ \cite{cheng2020cspn++} to enlarge propagation neighbourhoods. Recently, depth completion with lightweight ToF sensors has emerged. Compared with RGB-D cameras, lightweight ToF sensors (e.g., L5) are low-power and realistic \cite{li2022deltar,he2021towards}, but they only provide coarse depth distribution in each zone and have a low resolution (e.g., $8 \times 8$). DELTAR \cite{li2022deltar} is a framework that includes D-to-image cross attention to propagate depth distribution features into image features in a patch-wise way. Authors of DELTAR also notice the FOV difference between the ToF sensor and the RGB camera, and utilize self-attention \cite{vaswani2017attention} to propagate ToF features to global contexts. However, we notice that this self-attention is not enough and propose more effective cross-zone feature propagation modules to achieve better performance in outside-zone areas.

\subsection{Long-Range Dependency}  In computer vision, large receptive fields and long-range dependencies used to be acquired by stacking convolutional blocks \cite{lecun1989backpropagation}. Ever since the emergence of transformer blocks\cite{vaswani2017attention}, long-range dependency has been modeled by these blocks and proven to be effective in semantic segmentation \cite{xie2021segformer, guo2022cmt}, image restoration \cite{zamir2022restormer}, depth completion \cite{zhang2023completionformer}, BEV perception \cite{li2022bevformer}. These successful applications of transformer blocks \cite{vaswani2017attention} validate the necessity of long-range dependency modeling in deep neural networks \cite{park2020non}. Another way to achieve long-range dependency is to use convolutions with large kernels. Modern networks, including ResNet \cite{he2016deep}, DenseNet \cite{huang2017densely}, EfficientNet \cite{tan2019efficientnet}, etc., mostly use convolution of size $3 \times 3$. Yet, models with large kernels (e.g., $9 \times 9$) are also effective \cite{peng2017large} in image classification and localization. Recently, RepLKNet \cite{ding2022scaling} prove that kernel sizes as large as $31 \times 31$ can attain very large effective receptive fields (ERFs) \cite{luo2016understanding}. Based on these observations, we propose two modules that contain transformer-like blocks \cite{vaswani2017attention} and 
convolution with large kernels \cite{ding2022scaling}, respectively, to build long-range dependency for effective cross-zone feature propagation.
\section{Method}

In the task of ToF-based depth completion, we aim to infer a depth map $D \in \mathbb{R}^{H\times W \times 1}$ given RGB image $I \in \mathbb{R}^{H\times W \times 3}$ and a depth distribution map $D \in \mathbb{R}^{h\times w \times c_t}$, where $h << H$ and $w << W$ (e.g., $H=480, W=640, h=8, w=8$). Notice that $c_t$ can vary depending on device setups and implementation details. Here, we follow DELTAR \cite{li2022deltar} to sample 16 values for each zone based on collected mean and variance from the L5 measurements, i.e., $c_t = 16$. 

As discussed earlier, the performance drop in outside-zone areas can be mitigated by leveraging ToF information from zone areas. Thus, we aim to propagate features in zone areas, which contain depth information, into outside-zone areas. In this section, we first introduce the overall network architecture of our CFPNet, which is based on the DELTAR model. Then, we illustrate our newly proposed Direct-Attention-based Propagation Module (DAPM) and Large-Kernel-based Propagation Module (LKPM), both of which are designed for establishing long-range feature propagation for pixels in outside-zone areas from in-zone areas. Additionally, we empirically verify how to combine these two modules effectively to achieve the best performance. Lastly, we discuss the loss function used to train the model. 

\subsection{Network Architecture}
 
\begin{figure*}[t]
    \centering
    \includegraphics[width=1.0\textwidth]{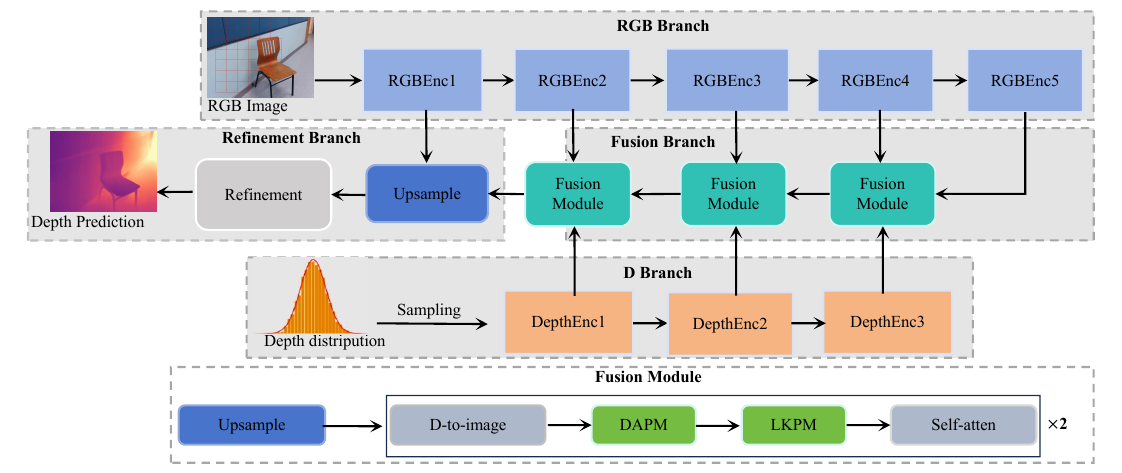}
    \caption{The architecture of our CFPNet. Our CFPNet takes RGB image and depth distribution from ToF sensors as inputs and outputs the depth completion prediction. Our newly proposed DAPM and LKPM are located in the fusion module and allow effective cross-zone feature propagation from zone areas to outside-zone areas. }
    \label{fig:network}
\end{figure*}

We build our CFPNet based on the DELTAR model, and add our DAPM and LKPM for more effective cross-zone feature propagation. As illustrated in Fig. \ref{fig:network}, our CFPNet is composed of four branches: RGB feature extraction branch (RGB Branch), depth distribution branch (D Branch), fusion branch, and refinement branch.

\textbf{RGB Branch.} We use a popular convolutional encoder, i.e., EfficientNetB5 \cite{tan2019efficientnet}, to extract RGB features at multiple resolutions. Consequently, five RGB feature maps are generated, which will be utilized in the following branches.

\textbf{D Branch.} Given ToF measurements, we first sample 16 depth values for each zone. In this way, each zone is considered as one point with sampled depth values as features. Then PointNet-like \cite{qi2017pointnet} structure is used to extract depth distribution features. Since the resolution of the zone area is already small, we do not conduct downsampling operations.

\textbf{Fusion Branch.} After acquiring RGB and depth distribution features, respectively, we aim to fuse these features effectively. This branch contains three fusion modules at different levels. Each fusion module is composed of the upsampling module, the D-to-image module \cite{li2022deltar}, our DAPM, our LKPM, and the self-attention module \cite{vaswani2017attention}. The upsampling module upsamples the low-resolution feature map and concatenates it with RGB features from higher resolution. After upsampling, D-to-image module would conduct cross-attention from depth distribution features to RGB features in a patchwise way. Concretely, we generate keys and values from depth distribution features and generate queries from RGB features. Thus, the RGB features can dynamically retrieve information from depth distribution features. After fusing these two types of features in the zone area, we utilize DAPM (Sec. \ref{sec:APM}) and LKPM (Sec. \ref{sec:LKPM}) to propagate features from zone areas to outside-zone areas. Lastly, a self-attention layer is added to blend the feature maps in a global context and make it slightly smoother. Note that apart from the upsampling module, the rest modules are applied alternatively twice.

\textbf{Refinement Branch.} Due to the large resolution of feature maps at the highest level, we do not conduct feature fusion here. Yet, we directly upsample the output from the fusion branch and apply a refinement module to generate the predicted depth map. The refinement module is the same as in DELTAR,  which is a mViT structure from Adabins \cite{bhat2021adabins}. This module predicts the depth in a weighted sum of adaptively predicted depth bins.

\subsection{Direct-Attention-based Propagation Module} 
\label{sec:APM}
We propose to utilize the long-range dependency modeling ability of transformer-like \cite{vaswani2017attention} structure to directly propagate features from zone areas to outside-zone areas. Compared with using self-attention \cite{li2022deltar}, our feature propagation avoids feature queries from outside-zone areas where ToF information does not exist. 

\begin{figure}[t]
    \centering
    \includegraphics[width=0.5\textwidth]{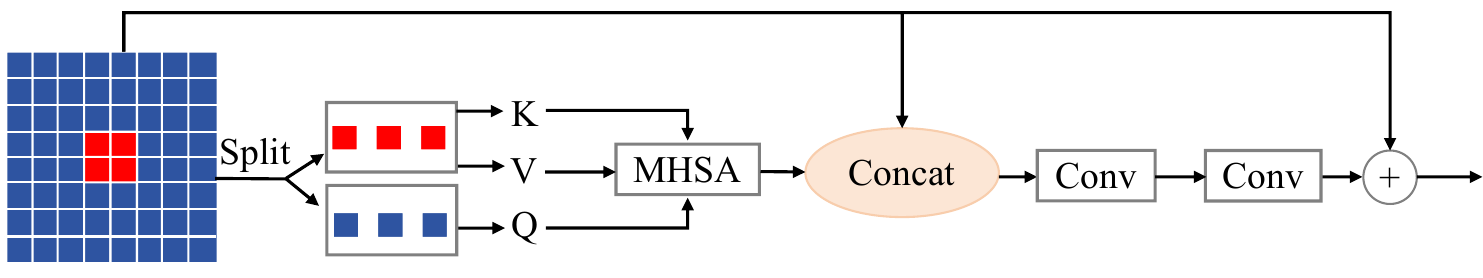}
    \caption{The pipeline of proposed DAPM. We conduct cross attention between pixels from zone areas and outside-zone areas. Additional convolution and skip connection are added to capture local contexts and promote propagation of gradients, respectively. }
    \label{fig:APM}
\end{figure}

Consider the toy example in Fig. \ref{fig:APM}. Given a feature map, we split the pixels into two groups depending on whether they are in or outside zone areas. This results in two sequences of tokens where each pixel is a token. Then, we leverage linear cross attention \cite{katharopoulos2020transformers} to propagate features efficiently. Specifically, we generate keys and values from zone areas, and queries from outside-zone areas. Then, Multi-Head Self-Attention (MHSA) \cite{vaswani2017attention} is used to calculate the queried values. Consequently, the features can be propagated to outside-zone areas dynamically. Notice that this operation has a global receptive field and is robust to the location of the zone areas in the image plane. Next, we concatenate the output from MHSA with the input and apply a convolution to restore the channel 
number as the input as in the D-to-image module. Additionally, we employ one more $3 \times 3$ convolution layer and skip connection  \cite{he2016deep} to capture local contexts with CNN in addition to transformer blocks \cite{wu2021cvt} and promote gradient propagation \cite{guo2022cmt}, respectively. Furthermore, we validate that the convolution layer and skip connection \cite{he2016deep} in the end are necessary in the ablation study (Table \ref{table:APM}). 

\subsection{Large-Kernel-based Propagation Module} 
\label{sec:LKPM}

\begin{figure}[t]
    \centering
    \includegraphics[width=0.5\textwidth]{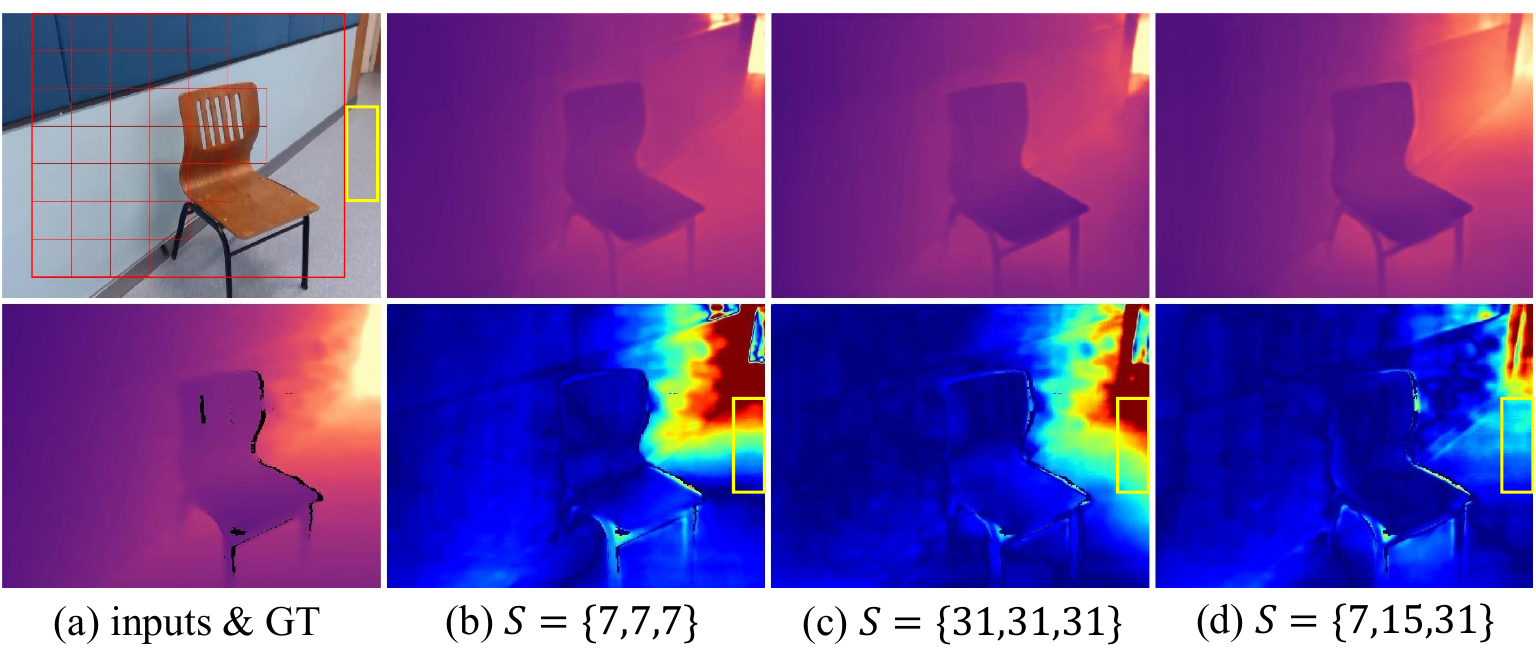}
    \caption{Qualitative results comparing different kernel designs in our LKPM on ZJU-L5 dataset. Compared with using only small or large kernel sizes, our adaptive kernel design achieves the best performance.}
    \label{fig:LK_comparisons}
\end{figure}

Convolution layers can also be adopted to perform feature propagation, and their receptive fields are not influenced by similarities among pixels as in self-attention \cite{bello2019attention}.  In this problem, we propose to use convolution layers of very large kernels (e.g., 31) for cross-zone feature propagation. The usage of large kernels is the large distance from zone areas to outside-zone areas, especially when large portions of zones are missing as in Fig. \ref{fig:LK_comparisons} (a). Furthermore, large kernels can have larger effective receptive fields \cite{luo2016understanding} compared with multiple small kernels \cite{ding2022scaling}. To this end, we introduce a large-kernel-based propagation module (LKPM) based on RepLKNet \cite{ding2022scaling} and ConvNeXt\cite{liu2022convnet}.

\begin{figure}[t]
    \centering
    \includegraphics[width=0.4\textwidth]{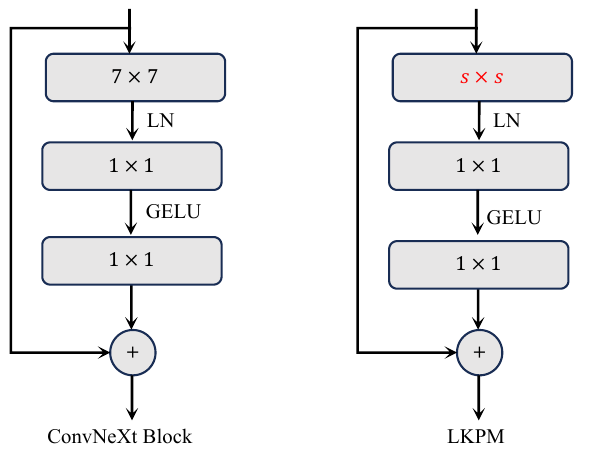}
    \caption{Designs of ConvNeXt Block and our LKPM. Different from ConvNext, we use a $s \times s$ convolution layer where $s$ could be as large as 31 instead of fixing $s$ as 7. Moreover, we adaptively set $s$ based on the resolution of feature maps. }
    \label{fig:LKPM}
\end{figure}

Fig. \ref{fig:LKPM} (b) depicts our proposed LKPM. It is composed of a depthwise convolution layer of size $s \times s$ (e.g., $s=31$), LayerNorm \cite{ba2016layer}, two $1 \times 1$ convolution layers and the GELU \cite{hendrycks2016gaussian} activation unit. Notably, the depthwise convolution with the same number of groups and channels allows the usage of convolution with large kernels \cite{ding2022scaling}. Compared with the ConvNeXt block \cite{liu2022convnet}, rather than using $7 \times 7$ convolution layers, we use convolution layers with much larger kernel sizes. The large kernel allows more long-range feature aggregation between pixels from zone areas and outside-zone areas. Furthermore, we heuristically set $s$ based on the resolution of current feature maps, as we find that changing the kernel size in LKPM adaptively is better than setting $s = 31$ throughout the used blocks \cite{ding2022scaling}. Specifically, given input of size $480 \times 640$, the fusion happens at three stages where the feature maps are of size $30 \times 40$, $60 \times 80$, $120 \times 160$. Thus, we empirically set $S=\{7,15,31\}$ for LKPM deployed in these stages where $S$ is the collection of $s$ used in three stages. 


A visual comparison between using ConvNeXt design \cite{liu2022convnet} ($S=\{7,7,7\}$), RepLKNet design \cite{ding2022scaling} ($S=\{31,31,31\}$), and our kernel design ($S=\{7,15,31\}$) are given in Fig. \ref{fig:LK_comparisons}. The error maps clearly validate that using merely small kernels \cite{liu2022convnet} is not likely to propagate the features from zone areas to outside-zone areas, especially when large portions of zones are missing. Also, using only large kernels would oversmooth the predicted depth map (faraway areas are estimated nearer). Our kernel design significantly reduces the depth prediction errors in outside-zone areas in the yellow rectangles. More discussions on kernel sizes are in the ablation studies (Sec. \ref{ablation}).

\subsection{Combining DAPM and LKPM}

We have presented two modules, DAPM and LKPM, to propagate features from zone areas to outside-zone areas. DAPM enjoys the power of attention and has a global receptive field \cite{vaswani2017attention}. On the other hand, LKPM benefits from the locality of CNN, though we use large kernels. Since CNN-Transformer-like combinations have been proven to be effective in image classification \cite{guo2022cmt}, depth completion \cite{zhang2023completionformer}, etc., we propose to combine them for the best performance. 

To this end, we test three ways of using DAPM and LKPM jointly. In model A, we apply DAPM first and then apply LKPM. In model B, we apply LKPM first and then apply DAPM. In model C, we apply LKPM and DAPM in a parallel way and fuse the two results with a summation. The rest of models A, B, and C are the same as in our baseline, which is the DELTAR\cite{li2022deltar} model. Table \ref{table:ablation} validates that model A would perform the best on the ZJU-L5 dataset \cite{li2022deltar}.

Furthermore, combining them would lead to more robust performance in various FOV difference setups. For the small FOV difference case as in the ZJU-L5 dataset, LKPM could be more effective than DAPM (see Table \ref{table:ablation}). However, in a large FOV difference scenario, DAPM can be more effective (see Table \ref{table:diff}) thanks to the global perception ability from attention mechanism \cite{vaswani2017attention}. Consequently, combining them is necessary to handle different FOV setups in real-world applications.

\subsection{Loss Function}

In order to train a depth completion network, commonly chosen loss functions can be L1 loss, L2 loss, BerHu loss \cite{owen2007robust}, Scale-Invariant (SI) loss \cite{eigen2014depth} , etc. We follow previous works \cite{bhat2021adabins}, and use a scaled version of SI loss for each sample:

$$
L(d,\tilde{d})=\alpha \sqrt{\frac{1}{N} \sum_{i=1}^{N} g_i^2-\frac{\lambda}{N^2}\left(\sum_{i=1}^{N} \tilde{g}_i\right)^2}
$$
where $d$ and $\tilde{d}$ is the ground truth depth and predicted depth for the sample, $N$ is the number of valid pixels, $g_i=\log \tilde{d}_i-\log d_i$. $\alpha$ and $\lambda$ are set to 10 and 0.85.

\section{Experiments}
In this section, we first describe datasets and evaluation metrics. Then, quantitative and qualitative results are provided to validate the remarkable
 performance of our CFPNet. Lastly, abundant ablation studies are given to verify the effectiveness of our proposed DAPM and LKPM.

\subsection{Datasets and Evaluation Metrics}

\textbf{NYU-sim} NYU-sim is a simulated dataset from NYUDepth-V2 \cite{silberman2012indoor}, containing RGB images, simulated ToF measurements, and groundtruth depth maps. The training and test sets contain 24k and 654 samples, the same as previous works \cite{lee2019big}. As for simulating the ToF signal, a Gaussian distribution is used to fit the depth histogram for each zone. Pixels whose depth is farther than the range the ToF sensor can measure are excluded during the histogram statistics.

\textbf{ZJU-L5.} DELTAR \cite{li2022deltar} provided a real-world test set, ZJU-L5, containing 527 samples from 15 different scenes. 

\textbf{Metrics.} We follow previous works \cite{bhat2021adabins} to report standard metrics for depth prediction, including mean absolute relative error (REL), root mean squared error (RMSE), average $\left(\log _{10}\right.$) error,  threshold accuracy $\left(\delta_i\right)$.


\subsection{Implementation Details}

We implement our CFPNet in Pytorch \cite{paszke2019pytorch}. For training, we use AdamW optimizer \cite{loshchilov2017decoupled} with one-cycle policy \cite{smith2019super} where maximum learning rate is of $3 \times 10^{-4}$. We train our CFPNet on four NVIDIA RTX 2080Ti GPUs with a batchsize of 16 on each GPU for 30 epochs, which takes around 12 hours. 

As for training and test protocol, we follow previous works to first train on the NYU-sim dataset and select the model with the best performance on the test set of the NYU-sim dataset. Then, we report the performance of this model on the ZJU-L5 dataset \cite{li2022deltar}.

\subsection{Quantitative Comparisons} 

Table \ref{table:quantitative} lists the performance of different methods including BTS \cite{lee2019big}, Adabins \cite{bhat2021adabins}, NLSPN \cite{park2020non}, PENet \cite{hu2021penet}, DELTAR \cite{li2022deltar}, our reproduced DELTAR* \cite{li2022deltar} using their most recent codebase, and our CFPNet, on ZJU-L5 dataset \cite{li2022deltar}. The first five lines of results are quoted from DELATR \cite{li2022deltar}. Overall, Table \ref{table:quantitative} shows that specifically designed lightweight-ToF-based depth completion (DC) methods, i.e., DELTAR \cite{li2022deltar} and our CFPNet, are more effective than previous depth estimation (DE) and depth completion (DC) methods.  

Note that DELTAR* is our baseline and has similar performance compared with DELTAR, except on the $\delta_1$ and RMSE metric. Thus, we mainly compare our CFPNet with DELTAR* regarding quantitative and qualitative results. Our CFPNet outperforms DELTAR* by a notable margin. For example, our CFPNet can decrease the REL by 0.024 and increase the $\delta_1$ by 0.021. This validates that our CFPNet can achieve SOTA performance in lightweight-ToF-based depth completion.

\begingroup
\setlength{\tabcolsep}{10pt} 

\begin{table}[htbp]
\resizebox{\columnwidth}{!}{
\centering
\begin{tabular}{cccccccc}
\hline Methods & type & $\delta_1 \uparrow$ & $\delta_2 \uparrow$ & $\delta_3 \uparrow$ & REL $\downarrow$ & RMSE $\downarrow$ & $\log _{10} \downarrow$ \\
\hline 
BTS \cite{lee2019big}          & DE & 0.739 & 0.914 & 0.964 & 0.174 & 0.523 & 0.079 \\
AdaBins \cite{bhat2021adabins} & DE & 0.770 & 0.926 & 0.970 & 0.160 & 0.494 & 0.073 \\
NLSPN \cite{park2020non}       & DC & 0.583 & 0.784 & 0.892 & 0.345 & 0.653 & 0.120 \\
PENet \cite{hu2021penet}       & DC & 0.807 & 0.914 & 0.954 & 0.161 & 0.498 & 0.065 \\
DELTAR \cite{li2022deltar}     & DC & 0.853 & 0.941 & \textbf{0.972} & 0.123 & 0.436 & 0.051 \\
DELTAR* \cite{li2022deltar}    & DC & 0.862 & 0.943 & 0.970 & 0.127 & 0.461 & 0.058 \\
CFPNet                           & DC & \textbf{0.883} & \textbf{0.949} & \textbf{0.972} & \textbf{0.103} & \textbf{0.431} & \textbf{0.047} \\
\hline
\end{tabular}}
\caption{Quantitative comparisons on the ZJU-L5 dataset. Compared with the most recent lightweight-ToF-based method DELTAR* (our reproduced baseline), we obtain large improvement on all metrics. Best performance is \textbf{bolded}.}
\label{table:quantitative}
\end{table}
\endgroup

\subsection{Qualitative Comparisons}

\begin{figure}[t]
    \centering
    \includegraphics[width=0.5\textwidth]{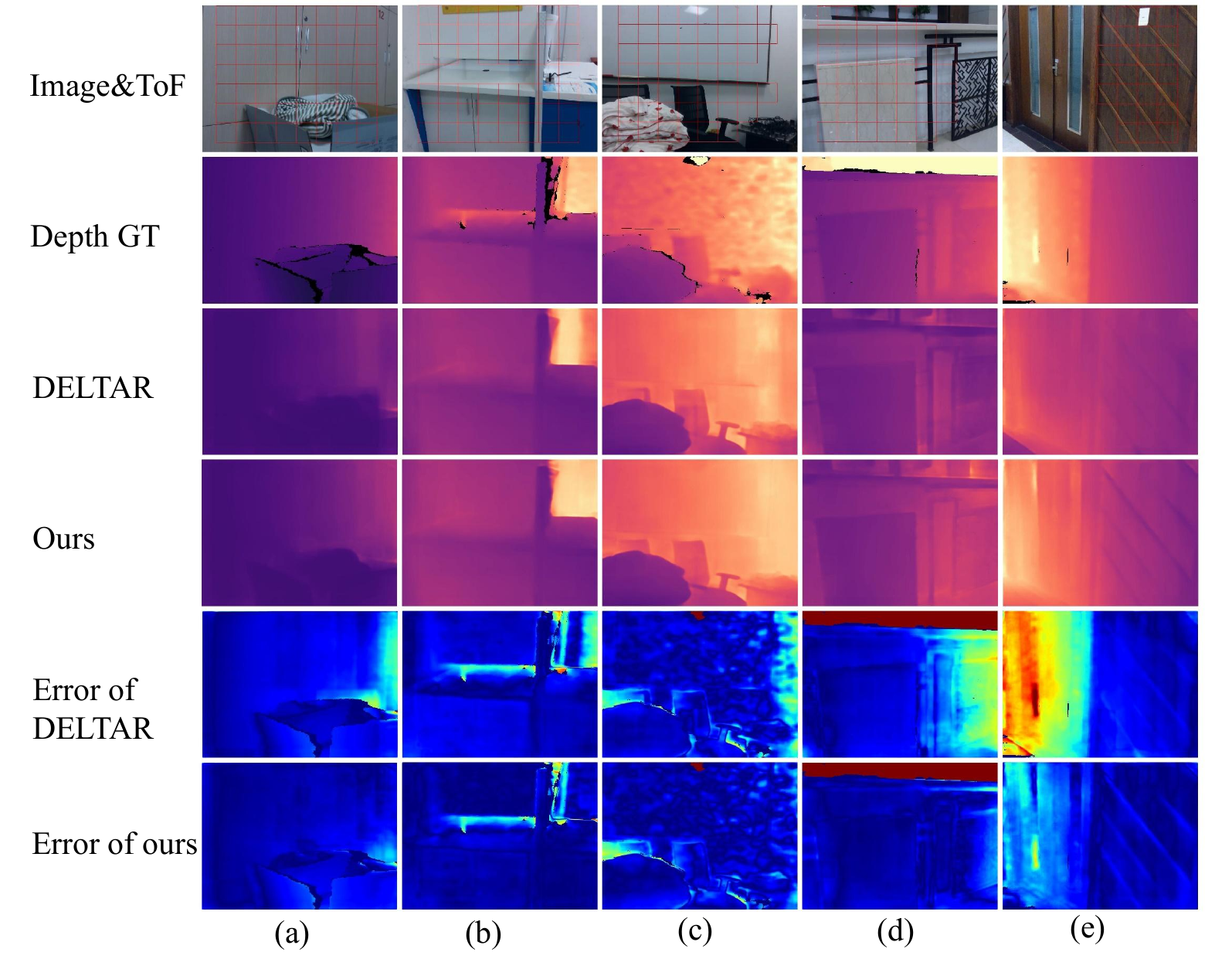}
    \caption{Qualitative comparisons between DELATR and our CFPNet on the ZJU-L5 dataset. Brighter color in error maps refers to larger errors. Errors where depth gt are missing are set to zero for visualization. Clearly, in outside-zone areas, errors are greatly reduced, and 3D properties such as planar smoothness are maintained.}
    \label{fig:qualitative}
\end{figure}

To validate that our CFPNet can improve depth completion performance in outside-zone areas, we provide visual comparisons between DELTAR* and our CFPNet in Fig. \ref{fig:qualitative}.

Obviously, our CFPNet can maintain the scene's structures better and restore more accurate depth in outside-zone regions. For example, in Fig. \ref{fig:qualitative} (a), (b) and (c), the prediction on the right side of the image preserves the continuity of the wall. In contrast, DELTAR* would generate quite different predictions for in-zone pixels and outside-zone pixels of the wall. Furthermore, in more challenging scenarios, e.g., Fig. \ref{fig:qualitative} (e), even if the majority of the ToF signal is lost on the left, our CFPNet can still recover the depth of the scene with higher precision. These visualizations validate that our CFPNet can effectively mitigate the performance drop in outside-zone areas. More qualitative results are provided in the supplementary material.

\subsection{Ablation studies}
\label{ablation}
To understand the impact of each proposed module, we conduct thorough ablation studies in this section. We first verify the improvement each module incurs and then examine the effectiveness of our designs in the module. Lastly, we discuss the differences between these two modules regarding application scenarios.

Table \ref{table:ablation} shows that LKPM and DAPM can increase $\delta_1$ by 0.019 and 0.010, respectively. They can also reduce RMSE by 0.022 and 0.014. We also test different combinations of DAPM and LKPM. Model A performs DAPM first and then LKPM, while model B is vice-versa. Model C conducts these two modules parallelly and sums the outputs from them. Table \ref{table:ablation} suggests that model A can achieve the best performance. Additionally, we propose model D, which is the same as model A, except we remove the self-attention layer \cite{vaswani2017attention}. The better performance and fewer parameters of model D compared with DELTAR* indicates that our proposed DAPM and LKPM are more effective and efficient than self-attention \cite{vaswani2017attention} in this task. Still, applying self-attention \cite{vaswani2017attention} as in model A leads to even better performance.

\begingroup
\setlength{\tabcolsep}{10pt} 

\begin{table}[htbp]
\resizebox{\columnwidth}{!}{
\centering
\begin{tabular}{ccccccccc}
\hline Methods  & $\delta_1 \uparrow$ & $\delta_2 \uparrow$ & $\delta_3 \uparrow$ & REL $\downarrow$ & RMSE $\downarrow$ & $\log _{10} \downarrow$ & Params(M) \\
\hline 
DELTAR* \cite{li2022deltar}     & 0.862 & 0.943 & 0.970 & 0.127 & 0.461 & 0.058 & 18.545($\pm0.000$)\\
+LKPM                           & 0.881 & 0.947 & 0.971 & 0.104 & 0.439 & 0.049 & 18.996($+0.451$)\\
+DAPM                            & 0.872 & 0.945 & 0.970 & 0.112 & 0.447 & 0.050 & 19.837($+1.292$)\\
\hline 
Model C                         & 0.875 & 0.946 & 0.971 & 0.107 & 0.436 &0.049 & 20.288($+1.743$) \\
Model B                         & 0.877 & 0.943 & 0.969 & \textbf{0.102} & 0.442 & 0.049 & 20.288($+1.743$)\\
Model A                         & \textbf{0.883} & \textbf{0.949} & \textbf{0.972} & 0.103 & \textbf{0.431} & \textbf{0.047} & 20.288($+1.743$)\\
Model D                         & 0.874 & 0.946 & 0.973 & 0.111 & 0.435 & 0.050 & 17.285($-1.260$)\\
\hline
\end{tabular}}
\caption{Ablation studies. We first reproduce DELTAR* as our baseline and add our DAPM or LKPM to investigate their individual effects. Then, we test different ways (Model A, B, C, D) to combine them. Both LKPM and DAPM can increase the overall performance, and Model A achieves the best performance.}
\label{table:ablation}
\end{table}
\endgroup

\textbf{Kernel Design in LKPM.} As mentioned above, larger kernels allow a larger effective receptive field \cite{ding2022scaling}, thus suitable for cross-zone feature propagation. Yet, we still need to investigate how large the kernel should be. To this end, we show the results of using different kernel sizes in LKPM in Table \ref{table:LKPM}. Overall, the performance can be improved with any configuration of LKPM, and setting the kernel sizes adaptively based on the resolution of feature maps (the fifth row) yields the best performance. However, using kernels with fixed size (7 or 31) or halving the kernel size in our design (the fourth row) can only lead to limited performance gain. Thus, these comparisons validate the effectiveness of our adaptive kernel design. Furthermore, we try adding an additional parallel convolution layer of $5 \times 5$ (the last row) as RePLKNet \cite{ding2022scaling}. Nevertheless, we find such a design is not useful in this task.

\begingroup
\setlength{\tabcolsep}{10pt} 

\begin{table}[htbp]
\resizebox{\columnwidth}{!}{
\centering
\begin{tabular}{cccccccc}
\hline Methods & Kernel sizes & $\delta_1 \uparrow$ & $\delta_2 \uparrow$ & $\delta_3 \uparrow$ & REL $\downarrow$ & RMSE $\downarrow$ & $\log _{10} \downarrow$ \\

\hline baseline  & -                   & 0.862  & 0.943 & 0.970 & 0.127 & 0.461 & 0.058 \\
+LKPM     & 7,7,7               & 0.870  & 0.945 & 0.969 & 0.116 & 0.448 & 0.053 \\
+LKPM     & 31,31,31            & 0.870   & 0.941	& 0.966	& 0.106	& 0.452	& 0.051 \\
+LKPM     & 3,7,15              & 0.867	& 0.943 & 0.969	& 0.110  & 0.458 & 0.052 \\
+LKPM     & 7,15,31             & \textbf{0.881} & \textbf{0.947} & \textbf{0.971} & \textbf{0.104} & \textbf{0.439} & \textbf{0.049} \\
+LKPM     & \{7,5\},\{15,5\},\{31,5\} & 0.878	& 0.945	& 0.969	& \textbf{0.104}	& 0.446 & 0.049 \\
\hline
\end{tabular}}
\caption{Ablation studies of kernel designs in LKPM. We investigate different combinations of kernel sizes used in LKPM in three levels. Using only small or large kernels could bring limited boost while our adaptive kernel design attains the best results.}
\label{table:LKPM}
\end{table}
\endgroup

\textbf{Skip Connection and Convolution in DAPM.}  As Fig. \ref{fig:APM} shows, DAPM includes cross-zone attention (CA) \cite{vaswani2017attention}, the final skip connection (SC), and the final convolution (Conv). We validate the necessity of each component in DAPM. Table \ref{table:APM} lists the performance of different models by injecting proposed components sequentially. Obviously, cross attention \cite{vaswani2017attention} improves the performance. Furthermore, adding the last skip connection and convolution yields the best performance in terms of overall metrics.

\begingroup
\setlength{\tabcolsep}{10pt} 

\begin{table}[htbp]
\resizebox{\columnwidth}{!}{
\centering
\begin{tabular}{cccccccc}
\hline Methods & $\delta_1 \uparrow$ & $\delta_2 \uparrow$ & $\delta_3 \uparrow$ & REL $\downarrow$ & RMSE $\downarrow$ & $\log _{10} \downarrow$ \\

\hline baseline     & 0.862  & 0.943 & \textbf{0.970} & 0.127 & 0.461 & 0.058 \\
+CA                 & 0.871	 & \textbf{0.945} & 0.968 & 0.111 & 0.458 & 0.053 \\
+CA+SC              & \textbf{0.873}	 & \textbf{0.945} & 0.967 & \textbf{0.107} & 0.459 & 0.052\\
+CA+SC+Conv         & 0.872	 & \textbf{0.945} & \textbf{0.970}  & 0.112 & \textbf{0.447} & \textbf{0.050} \\
\hline
\end{tabular}}
\caption{Ablation studies on components in DAPM. Though using cross-attention only could already bring some improvements, adding skip connection and convolution after attention increases the prediction accuracy steadily.}
\label{table:APM}
\end{table}
\endgroup

\begin{figure}[t]
    \centering
    \includegraphics[width=0.5\textwidth]{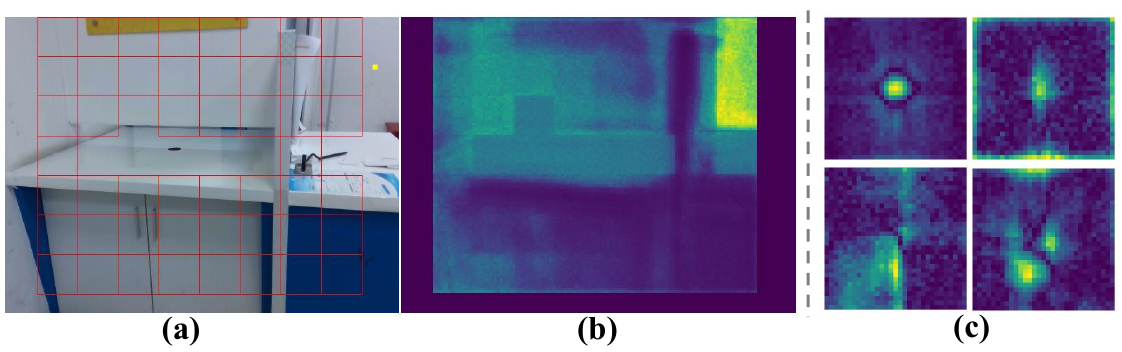}
    \caption{Visualization of learned attention map and kernel weights from our DAPM and LKPM. These results explain why DAPM and LKPM are capable of effectively propagating features from zone area to outside-zone area.}
    \label{fig:learned_weights}
\end{figure}

\textbf{Learned Attention Map and Large Kernel Weights.} In Fig.\ref{fig:learned_weights}, we show the attention map (b) for the selected yellow pixel in (a). We not only obtain higher attention weight in more relevant pixels but also avoid feature aggregation from outside-zone areas where ToF feature does not exist. In Fig. \ref{fig:learned_weights} (c), we show some of the learned large kernel weights which indicate that our large kernel can indeed propagate information in a broad context. These visualizations verify that our proposed DAPM and LKPM indeed allow effective feature propagation, 
 
\textbf{Discussion on DAPM and LKPM.} DAPM and LKPM can both improve the overall depth completion performance, and LKPM can lead to larger improvements. We argue that the small portion of outside-zone areas in the image limits the advantage of DAPM, which is designed to handle any FOV difference. To this end, we provide evidence that DAPM can be more useful in cases of a large FOV gap between the camera and the ToF sensor.




\begingroup
\setlength{\tabcolsep}{10pt} 

\begin{table}[htbp]
\resizebox{\columnwidth}{!}{
\centering
\begin{tabular}{ccccccc}
\hline Methods & $\delta_1 \uparrow$ & $\delta_2 \uparrow$ & $\delta_3 \uparrow$ & REL $\downarrow$ & RMSE $\downarrow$ & Time(ms) $\downarrow$ \\

\hline 
baseline        & 0.578  & 0.793  & 0.892 & 0.398 & 0.780  & 37.299 ($\pm0.000$)\\
+LKPM           & 0.607  & 0.833  & 0.919 & 0.321 &  0.651  & 40.366($+3.067$)  \\
+DAPM            & \textbf{0.701}  & 0.881  & 0.945 & 0.219 &  0.551  & 47.285($+9.986$)\\
+DAPM+LKPM       & 0.687  & \textbf{0.891}  & \textbf{0.952} & \textbf{0.213} &  \textbf{0.540} & 50.846($+13.547$)\\
+LKPM61            & 0.618 & 0.828 & 0.913 & 0.326 &0.629 & 45.271($+7.972$)\\
+LKPM101           & 0.638 & 0.845 & 0.921 & 0.302 & 0.612 &  58.821($+21.522$) \\
\hline
\end{tabular}}
\caption{Experimental results on ZJU-L5 where we simulate the condition that ToF is of resolution $2 \times 2$, i.e., large FOV difference between cameras and ToF sensors. In this scenario, DAPM can improve the performance more than LKPM.}
\label{table:diff}
\end{table}
\endgroup

The FOV difference between the camera and the ToF sensor in ZJU-L5 dataset \cite{li2022deltar} is small ($45 \degree \times 45 \degree$ versus $55\degree \times 43\degree$). Thus, when we conduct feature fusion at the level where the feature map is of size $120 \times 160$, ToF can cover regions of size $128 \times 128$, and each zone corresponds to $16 \times 16$. This is enough to cover the color image vertically but not horizontally. Assuming that the $128 \times 128$ area lies in the center of the color image, the horizontal distance from the left border of the image to the left border of the zone area is $(160-128)/2 = 16$ pixels. This is why the kernel size of 31 in LKPM is enough to handle outside-zone areas even if some zones are missing. However, if we simulate that the zone area is of resolution $2 \times 2$, which is a case of large FOV difference, this horizontal distance becomes $(160-2 \times 16)/2 = 64$ pixels. Then LKPM should bring less performance gain than DAPM in this situation due to its limited perceptive field. This is validated by results from Table \ref{table:diff}. More concretely, using DAPM yields more significant performance boost than using LKPM (REL is reduced by 45.0 \% and 19.3\%, respectively). Moreover, combining them as in model A still enjoys the advantage DAPM provides. Additionally, since kernel sizes in LKPM are specifically designed for $8\times8$ case, we provide two more results (LKPM61 and LKPM101) that use a maximum kernel size of 61 and 101. As the kernel size in LKPM increases, the latency increases significantly while the performance boost is limited.

\begin{figure}[t]
    \centering
    \includegraphics[width=0.5\textwidth]{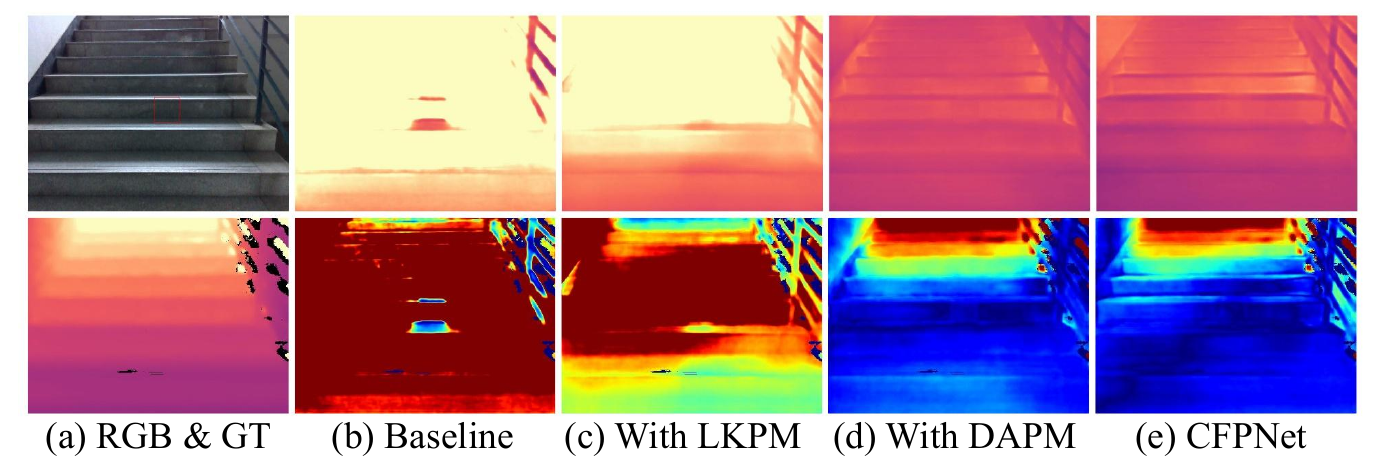}
    \caption{Visual comparisons under the condition where the ToF signal is of resolution $2 \times 2$ and only one zone is valid. This extreme case validates the superior performance of DAPM against LKPM when a large FOV difference between sensors exists.}
    \label{fig:diff}
\end{figure}

A visual comparison is also provided in Fig. \ref{fig:diff}. In this extreme case where we simulate that the ToF is of resolution $2 \times 2$ and only one zone is valid, the baseline method or using only the LKPM cannot give a reasonable prediction. However, the potential of DAPM is greatly excavated in this scenario. Moreover, combining them can preserve the benefits DAPM brings. More visualization results are in the supplementary material.

From Table \ref{table:ablation} and Table \ref{table:diff}, we find: (1) LKPM can be efficient (only increase the number of parameters by 0.451M) and effective on ZJU-L5 where the FOV difference is small. (2) DAPM can be more useful when the FOV difference is large (e.g., a more lightweight ToF sensor VL53L3CX \cite{VL53L3CX} has a FOV of $25\degree$), though at the cost of more parameters (1.292M). Still, combining them leads to more robust performance in different conditions.

\section{Conclusion and Future Work}
This paper proposes CFPNet containing two novel modules to tackle the FOV difference in RGB-guided lightweight-ToF-based depth completion by effective cross-zone feature propagation. The direct-attention-based feature propagation module attains direct feature acquisition via the attention mechanism. The large-kernel-based feature propagation module utilizes convolution layers of large kernels to achieve a large receptive field. Besides, we thoroughly discuss their differences and application scenarios. Extensive experiments demonstrate that our CFPNet achieves SOTA performance on the public dataset. However, one restriction is that due to the limited sensing range of the lightweight ToF sensor, depth completion performance in faraway regions is poor. 

As for future work, though we have gained a significant performance boost, this work is about ToF-based depth completion. There exists large room for ToF-based applications, including SLAM \cite{liu2023multi}, Nerf \cite{attal2021torf}, AR \cite{sun2023consistent}, etc. It is promising to utilize the low cost and energy property of L5 to conduct more meaningful downstream tasks. 

{
    \small
    \bibliographystyle{ieeenat_fullname}
    \bibliography{main}
}

\clearpage

\setcounter{page}{1}

\maketitlesupplementary

\begin{figure*}[t]
\includegraphics[width=1.0\textwidth]{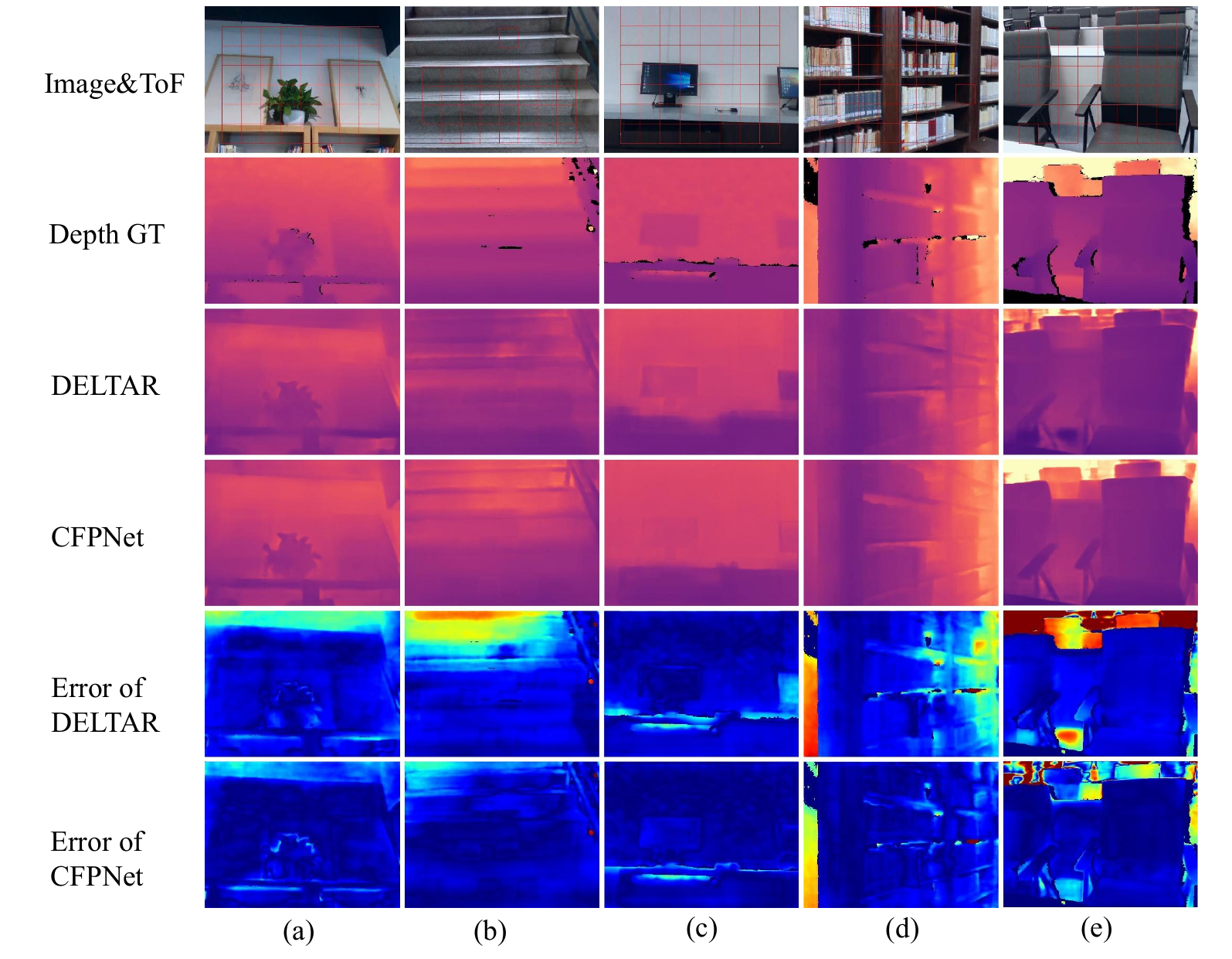} 
\captionof{figure}{More qualitative comparisons on the ZJU-L5 dataset. Our CFPNet achieve lower errors in outside-zone areas.}
\label{fig:qualitative_3}
\end{figure*}

\begin{figure*}[t]
\includegraphics[width=1.0\textwidth]{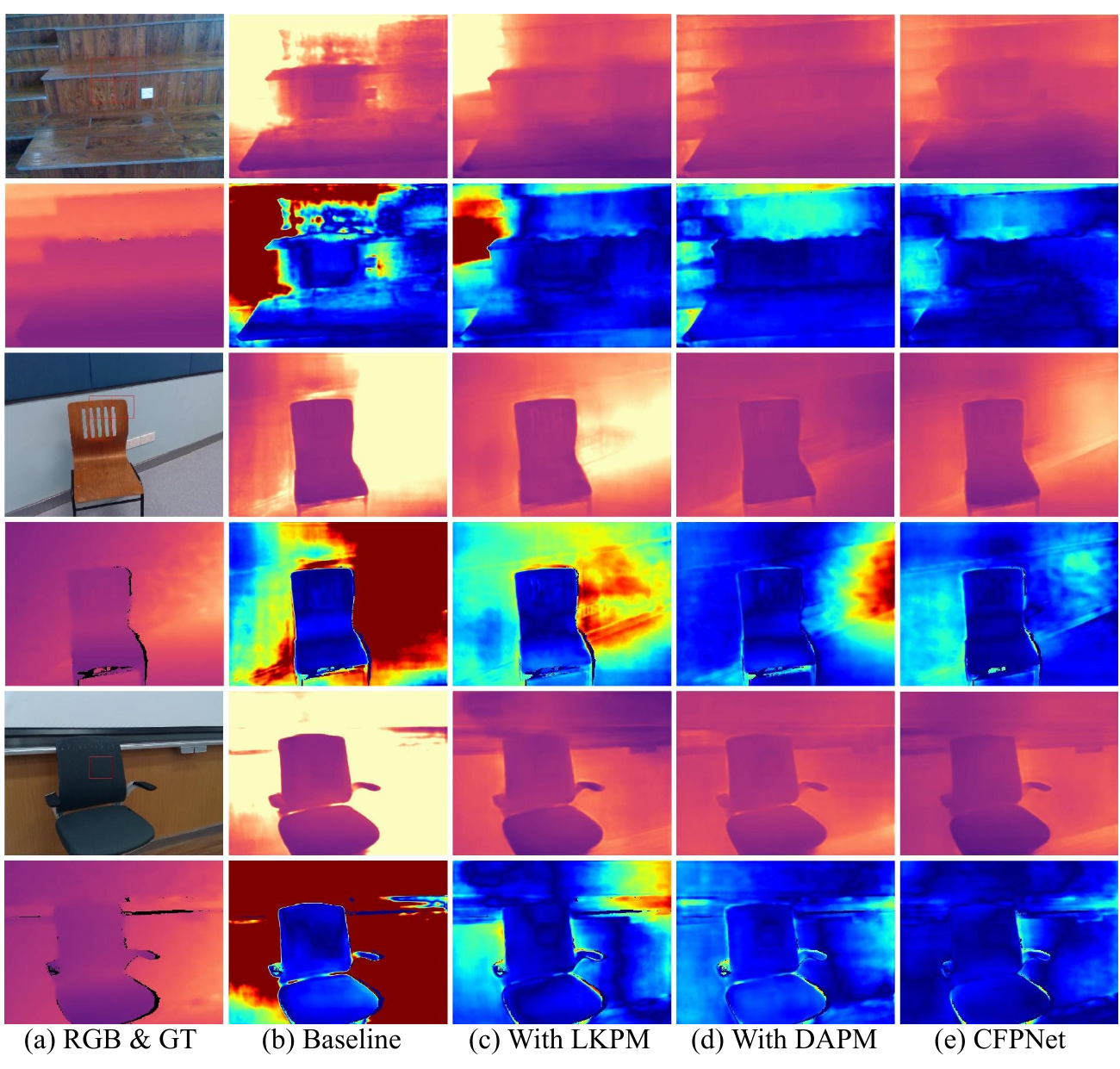} 
\captionof{figure}{More qualitative comparisons on the ZJU-L5 dataset where we simulate the ToF signal is of resolution $2 \times 2$. In this simulated large-FOV-difference case, our CFPNet can obtain superior performance compared with our baseline method \cite{li2022deltar}. Moreover, the benefits of DAPM are greater than those of LKPM in this scenario.}
\label{fig:qualitative_4}
\end{figure*}

\begin{strip}

In this supplementary material, we provide additional visual comparisons. More results comparing our CFPNet and DELTAR \cite{li2022deltar} are given in Fig. \ref{fig:qualitative_2} and \ref{fig:qualitative_3}. Furthermore, we offer additional results to demonstrate the superior depth completion performance of our CFPNet and the advantage of our direct-attention-based propagation module (DAPM) over large-kernel-based propagation module (LKPM) when the ToF signal is of resolution $2 \times 2$ in Fig. \ref{fig:qualitative_4}.

\includegraphics[width=1.0\textwidth]{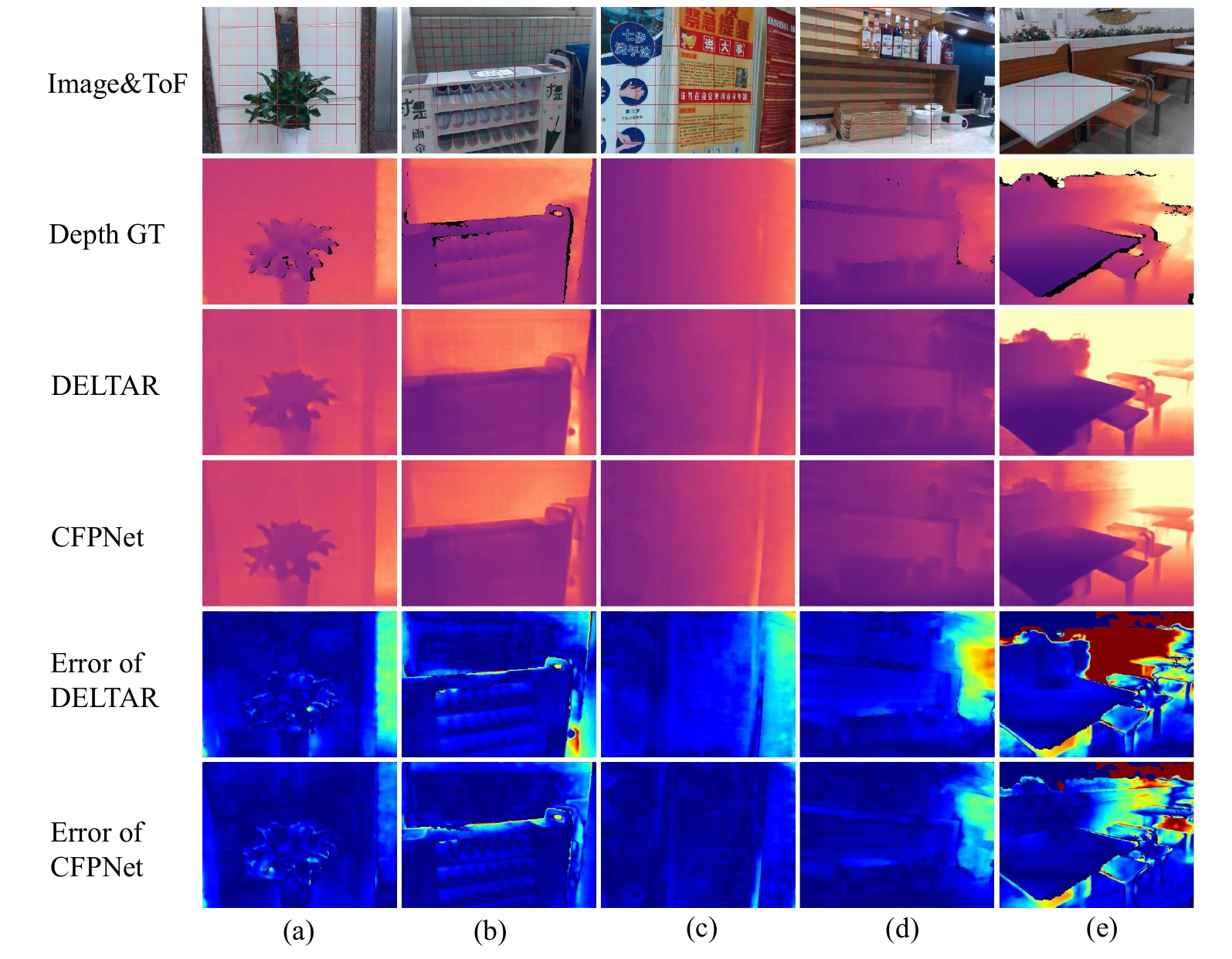} 
\captionof{figure}{More qualitative comparisons on the ZJU-L5 dataset. Our CFPNet achieve lower errors in outside-zone areas.}
\label{fig:qualitative_2}

\end{strip}

\end{document}